%% file: egbib.tex
\documentclass[runningheads]{llncs}
\usepackage{graphicx}
\usepackage{comment}
\usepackage{amsmath,amssymb} 
\usepackage{color}
\usepackage{subcaption}
\usepackage{mwe}

\usepackage{tabularx}
\usepackage[width=122mm,left=12mm,paperwidth=146mm,height=193mm,top=12mm,paperheight=217mm]{geometry}

\usepackage[vlined,ruled]{algorithm2e}

\begin{document}
\pagestyle{headings}
\mainmatter
\def\ECCVSubNumber{4372}  

\title{Exposing Deep-faked Videos by Anomalous Co-motion Pattern Detection}


\titlerunning{Abbreviated paper title}
%
\author{Gengxing Wang \and
Jiahuan Zhou\and
Ying Wu}
\authorrunning{G. Wang et al.}
%
\institute{Northwestern University
}

\maketitle
\vspace{-.5cm}

\begin{abstract}
Recent deep learning based video synthesis approaches, in particular with applications that can forge identities such as ``DeepFake'', have raised great security concerns.
Therefore, corresponding deep forensic methods are proposed to tackle this problem. 
However, existing methods are either based on unexplainable deep networks which greatly degrades the principal interpretability factor to media forensic, or rely on fragile image statistics such as noise pattern, which in real-world scenarios can be easily deteriorated by data compression. 
In this paper, we propose an fully-interpretable video forensic method that is designed specifically to expose deep-faked videos. 
To enhance generalizability on videos with various content, we model the temporal motion of multiple specific spatial locations in the videos to extract a robust and reliable representation, called \textit{co-motion pattern}. 
Such kind of conjoint pattern is mined across local motion features which is independent of the video contents so that the instance-wise variation can also be largely alleviated. 
More importantly, our proposed co-motion pattern possesses both superior interpretability and sufficient robustness against data compression for deep-faked videos. 
We conduct extensive experiments to empirically demonstrate the superiority and effectiveness of our approach under both classification and anomaly detection evaluation settings against the state-of-the-art deep forensic methods. 

\end{abstract}

\input{Sec_Intro}

\input{Sec_RW}

\input{Sec_Tech}

\input{Sec_Exp}

\vspace{-0.3cm}

\section{Conclusion $\&$ Future Work}

\vspace{-0.3cm}
In this work, we propose a novel co-motion pattern, a second-order local motion descriptor in order to detect whether the video is deep-faked. Our method is fully interpretable and pretty robust to slight variations such as video compression and noises. We have achieved superior performance on the latest datasets under classification and anomaly detection settings, and have comprehensively evaluated various characteristics of our method including robustness and generalizability. In the future, an interesting direction is to investigate whether a more accurate motion estimation can be achieved as well as how temporal information can be integrated within our method. 

\clearpage

\bibliographystyle{splncs04}
\bibliography{egbib}
\end{document}

%% file: Sec_Intro.tex
\section{Introduction}
Media forensic, referring to judge the authenticity, detect potentially manipulated region and reason its decision of the given images/videos, plays an important role in real life to prevent media data from being edited and utilized for malicious purposes, e.g., spreading fake news~\cite{FakeNews,WorldLeader}. Unlike traditional forgery methods (e.g., copy-move and slicing) which can falsify the original content with low cost but are also easily observable, the development of deep generative models such as generative adversarial net (GAN)~\cite{GAN} makes the boundary between realness and forgery more blurred than ever, as deep models are capable of learning the distribution from real-world data so well. In this paper, among all the forensic-related tasks, we focus on exposing forged videos produced by face swapping and manipulation applications~\cite{FastFaceSwap,DVP,F2F,FSGAN,MakeAFace,NeuralTexture}. These methods, while initially designed for entertainment purposes, have gradually become uncontrollable in particular when the face of celebrities, who possess greater social impact such as Obama~\cite{obama}, can be misused at no cost, leading to pernicious influence.

\begin{figure}[t!]
    \centering
    \includegraphics[width=0.95\textwidth, height=0.51\textwidth]{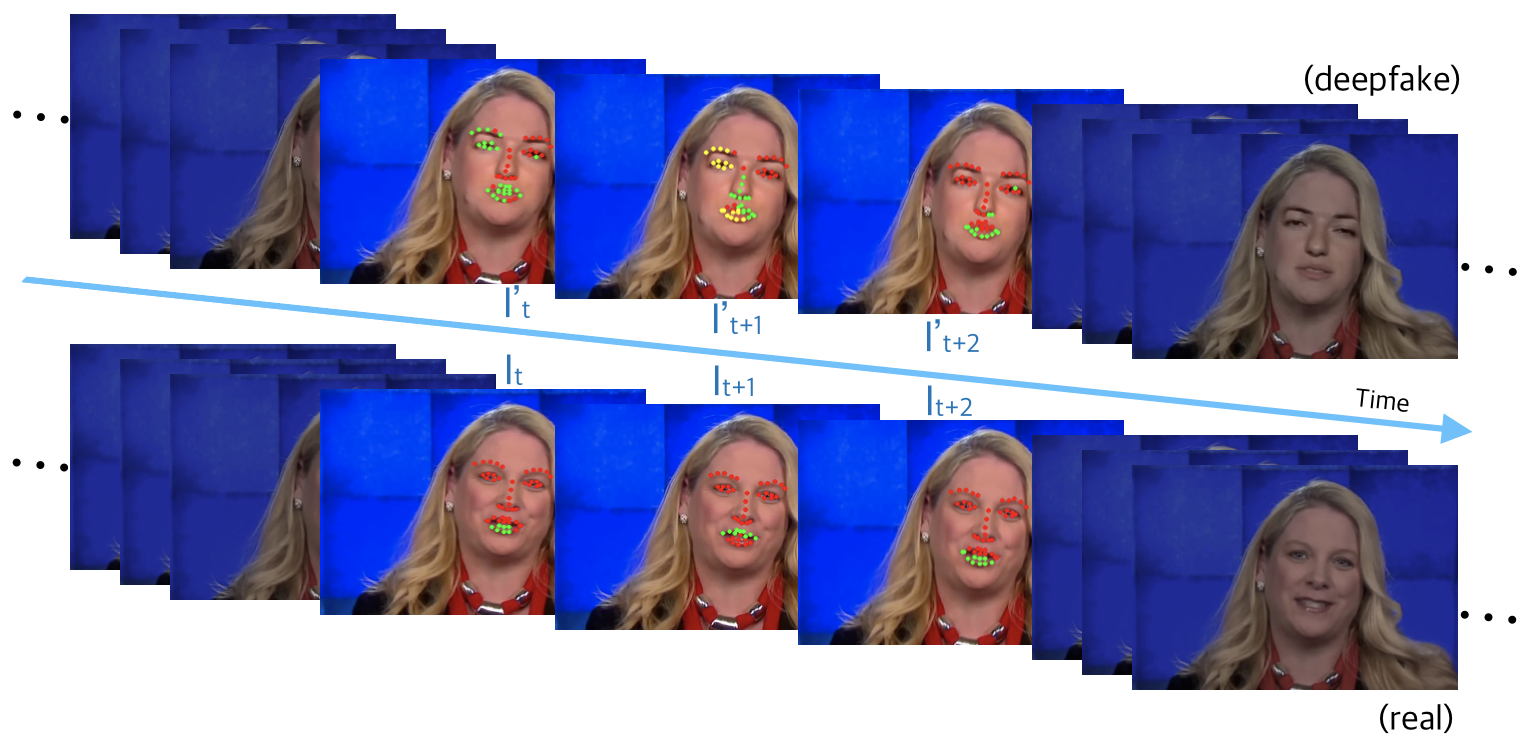}
    \caption{Example of motion analysis results by our method. \textbf{Landmarks} with the same color are considered having analogous motion patterns, which are consistent with facial structure in real videos but not in deep-faked videos. We compactly model such patterns and utilize them to determine the authenticity of given videos.}
    \label{fig:clear_comparison}
\end{figure}

Traditional forensic methods focusing on detecting specific traces remained ineluctably during the editing (e.g., inconsistency in re-sampling~\cite{Resampling}, shadowing~\cite{shadow}, reflection~\cite{Reflection}, compression quality~\cite{CompressionQuality} and noise pattern~\cite{Noise}) fail to tackle the indistinguishable DNN-generated images/videos due to the powerful generative ability of existing deep models. 
Therefore, the demand for forensic approaches explicitly against deep-faked videos is increasing. 
Existing deep forensic models can be readily categorized into three branches including real-forged binary classification-based methods~\cite{XRay,TwoStep,RCNN,MesoNet}, anomaly image statistics detection based approaches~\cite{ColorComponent,FaceArtifict,PRNU,Unmasking,AttributeGAN} and high-level information driven cases~\cite{headpose,exposelandmark,blinking}. 
However, no matter which kind of methods, their success heavily relies on a high-quality, uncompressed and well-labeled forensic dataset to facilitate the learning. Once the given data are compressed or in low-resolution, their performance is inevitably affected. More importantly, these end-to-end deep forensic methods are completely unexplainable, no explicit reason can be provided by these methods to justify based on what a real or fake decision is made.

To overcome the aforementioned issues, in this paper, we propose a video forensic method based on motion features to explicitly against deep-faked videos. Our method aims to model the conjoint patterns of local motion features from real videos, and consequently spot the abnormality of forged videos by comparing the extracted motion pattern against the real ones. To do so, we first estimate motion features of keypoints that are commonly shared across deep-faked videos. In order to enhance the generalizability of obtained motion features as well as eliminate noises introduced by inaccurate estimation results, we divide motion features into various groups which are further reformed into a correlation matrix as a more compact frame-wise representation. Then a sequence of correlation matrices are calculated from each video, with each weighted by the grouping performance to form the co-motion pattern which describes the local motion consistency and correlation of the whole video. In general, co-motion patterns collected from real videos obey the movement pattern of facial structures and are homogeneous with each other regardless of the video content variation, while it becomes less associated across fake videos. 

To sum up, our contributions are four-fold: (1) We propose co-motion pattern, a descriptor of consecutive image pairs that can be used to effectively describe local motion consistency and correlation. (2) The proposed co-motion pattern is being entirely explainable, robust to video compression/pixel noises and generalizes well. (3) We conduct experiments under both classification and anomaly detection settings, showing that the co-motion pattern is able to accurately reveal the motion-consistency level of given videos. (4) We also evaluate our method on datasets with different quality and forgery methods, with the intention to demonstrate the robustness and transferability of our method.

\begin{figure}[t!]
    \centering
    \includegraphics[width=\textwidth, height=0.45\textwidth]{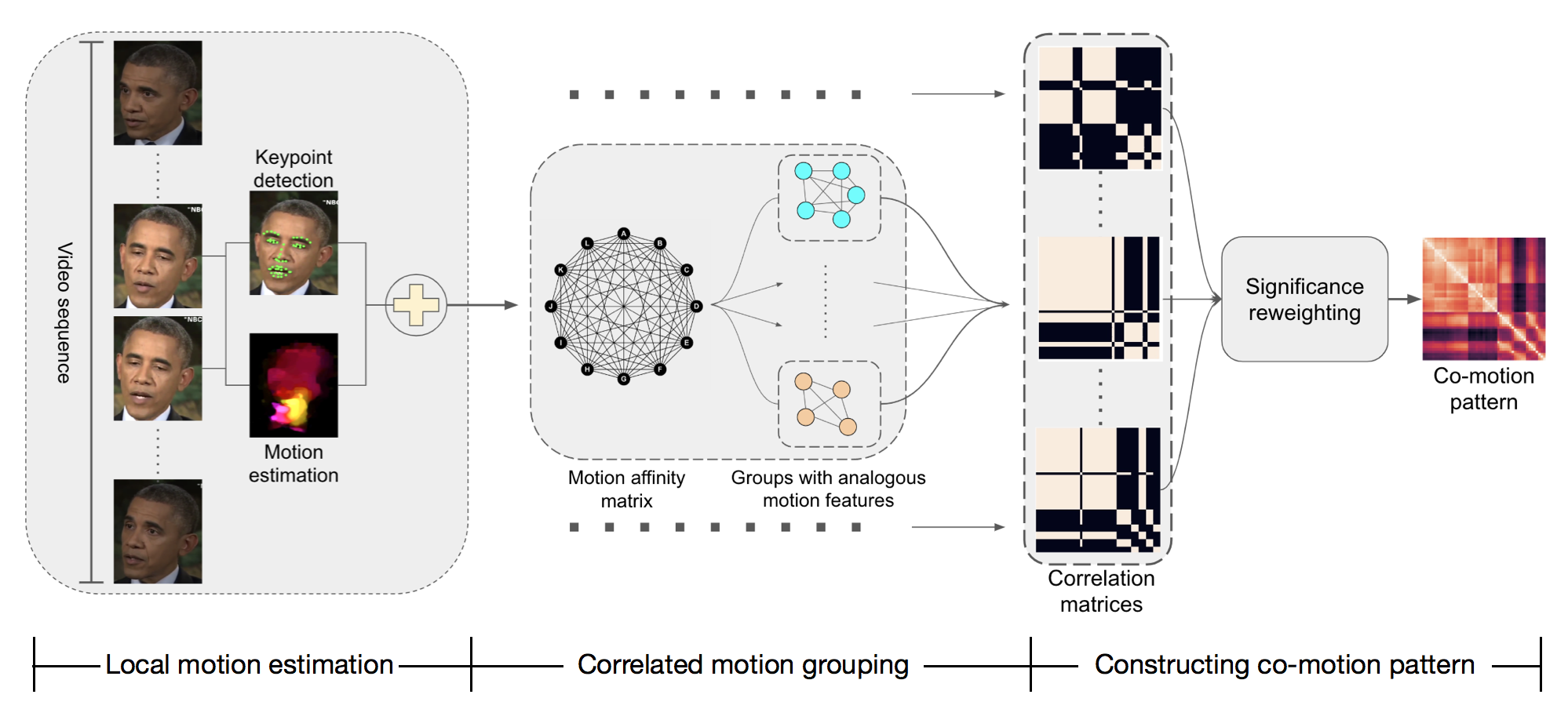}
    \caption{The pipeline of our proposed co-motion pattern extraction method. As illustrated, we firstly estimate the motion of corresponding keypoints, which are then to be grouped for analysis. On top of that, we construct co-motion pattern as a compact representation to describe the relationship between motion features. }
    \label{fig:framework}
\end{figure}

%% file: Sec_RW.tex
\vspace{-0.4cm}
\section{Related Work}
\vspace{-0.25cm}
\subsection{Face Forgery by Media Manipulation}
\vspace{-0.15cm}
First of all, we review relevant human face forgery methods. Traditionally, methods such as copy-move and slicing, if employed for face swapping tasks, can hardly produce convincing result due to the inconsistency caused by image quality~\cite{Resampling,quantization,jpeg_ghosts}, lighting changing~\cite{lighting,complex_lighting} and noise patterns~\cite{Noise,estimate_noise} between the tampered face region and other regions. With the rapid expeditious development of deep generative models~\cite{GAN}, the quality of generated images has significantly improved. The success of ProGAN~\cite{pggan} makes visually determining the authenticity of generated images pretty challenging if only focusing on the face region. Furthermore, the artifacts remained in boundary regions whose corresponding distribution in training datasets are relatively disperse are also progressively eliminated by \cite{StyleGANV1,StyleGANV2,glow,BigGAN}. Although these methods have demonstrated appealing generating capability, they do not focus on a certain identity but generate faces with random input. 

\begin{figure}
    \centering
    \includegraphics[width=0.8\textwidth, height=0.45\textwidth]{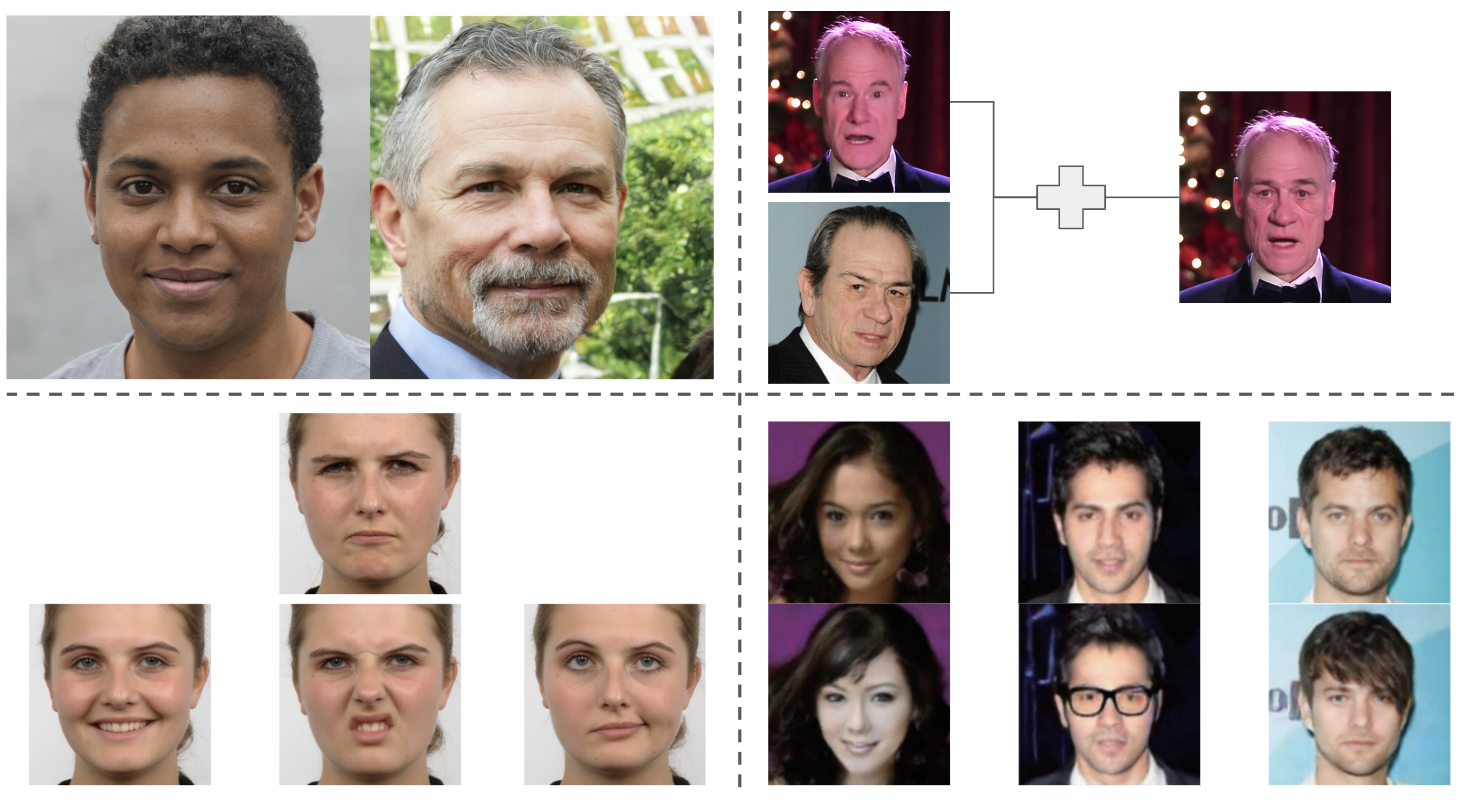}
    \caption{Samples to illustrate what ``Deepfake'' is. Top left~\cite{StyleGANV2}: high fidelity generated faces. Top right~\cite{jim}: face swapping. Bottom left~\cite{MakeAFace}: face expression manipulation, original image on top and expression manipulated on bottom. Bottom right~\cite{MulGAN}: face attribute editing, original images on top and edited on bottom. }
    \label{fig:deepfake_samples}
\end{figure}

Currently, the capability of deep neural networks has also been exploited for human-related tasks such as face swapping~\cite{deepfake,faceswap,FastFaceSwap,F2F,NeuralTexture,FSNET,FSGAN,DeformAE}, face expression manipulation~\cite{MakeAFace,F2F,x2face,NFE} and facial attribute editing~\cite{NFE,AttGAN,DA_Face_M,SMIT,MulGAN} majorly for entertainment purposes at the initial stage (samples of deep-faked face data are shown in Fig.~\ref{fig:deepfake_samples}.). However, since the face swapping methods in particular have already been misused for commercial purposes, homologous techniques should be studied and devised as prevention measures before it causing irreparable adverse influence. 

\vspace{-15pt}
\vspace{-15pt}

\subsection{Deep-faked Manipulation Detection}
While media forensic has been a long existing field, the countermeasures against deep-faked images and videos are scarce. As we mentioned earlier, existing methods can be categorized into three genres, respectively by utilizing a deep neural network~\cite{XRay,FaceForensics,RCNN,MesoNet,TwoStep,OFCNN,Incremental,DetectF2F,OpenWorld}, by exploiting the unnatural low-level statistics and by detecting the abnormality of high-level information. In the very first category, it has been usually considered as a binary classification problem where a classifier is constructed to learn the boundary between original and manipulated data via hand-crafted or deep features. As one of the earliest works in this branch, \cite{MesoNet} employs an Inception~\cite{Inception} with proper architecture improvement to directly classify each original or edited frame. Later, in order to consider the intra-frame correlation, \cite{RCNN} constructed a recurrent convolutional neural network that learns from temporal sequences. Due to the variety of video content and the characteristics of neural network, a sufficiently large dataset is required. To overcome this problem, \cite{OFCNN} attempted using the optical flow as input to train a neural network. While high classification accuracy achieved, since the features learned directly by neural networks yet to be fully comprehended, the decision of whether the input data has been manipulated cannot be appropriately elucidated. 

Regarding the second category, \cite{Unmasking,PRNU,AttributeGAN,CameraFingerprint} have all utilized the characteristics that the current deep generated images can barely learn the natural noise carried with untampered images, hence using the noise pattern for authentication. In \cite{ColorComponent}, the diminutive difference of color components between original and manipulated images for classification. While effective, these methods are also exceedingly susceptible to the quality of dataset. Our method lies in the third category and is constructed based upon high-level information~\cite{headpose,exposelandmark}, which are generally being more explainable and robust to the miniature pixel change introduced by compression or noise. Furthermore, as co-motion pattern is derived by second-order statistics, it is being more robust than ~\cite{headpose,exposelandmark} to instance-wise variation.

%% file: Sec_Tech.tex
\section{Methodology}
In this section, we elaborate on the details of our proposed video forensic method based on co-motion pattern extraction from videos and the overall pipeline of our method is illustrated in Fig.~2. Firstly, we obtain aligned local motion feature describing the movement of specific keypoints from the input videos (Sect.~\ref{sect:LME}). To eliminate the instance-wise deviation, we then design high-order patterns among the extracted local motion features. Subsequently, we demonstrate how to construct co-motion patterns that describe the motion consistency over each video, as well as its usage altogether in Sect.~\ref{sect:CMP}. 

\subsection{Local Motion Estimation}
\label{sect:LME}
The fundamental of constructing co-motion pattern is to extract local motion features firstly. Since each co-motion pattern is comprised by multiple independent correlation matrices (explained in Sect.~\ref{sect:CMP}), we expound on how to obtain local motion features from two consecutive frames in this section first. 

Denote a pixel on image $I$ with coordinate $(x, y)$ at time $t$ as $I(x, y, t)$, according to brightness constancy assumption, we have~\cite{HS,opticalflow}:
\begin{equation}
    I(x, y, t) = I(x + \Delta x, y + \Delta x, t + \Delta t)
\end{equation}
where $\Delta x, \Delta y$ and $\Delta t$ denote the displacements on $\mathbb{R}^3$ respectively. $\Delta t$ is usually 1 to denote two consecutive frames. This leads to the optical flow constraint:
\begin{equation}
    \frac{\partial I}{\partial x} \Delta x + \frac{\partial I}{\partial y} \Delta y + \frac{\partial I}{\partial t} = 0
\end{equation}
However, such a hard constraint can lead motion estimation result to be sensitive to even slight changes in brightness, and therefore gradient constancy assumption is proposed~\cite{gradient,opticalflow}:
\begin{equation}
    \nabla I(x, y, t) = \nabla I(x + \Delta x, y + \Delta y, t + 1)
\end{equation}
where
\begin{equation}
    \nabla = (\partial x, \partial y)^\intercal
\end{equation}
Based on above constraints, the objective function can be formulated as:
\begin{equation}
    \underset{\Delta x, \Delta y}{\min} E_{total}(\Delta x, \Delta y) = E_{brightness} + \alpha E_{smoothness}
\end{equation}
where:
\begin{equation}
\begin{split}
E_{brightness} = \iint & \psi(I(x, y, t) - I(x + \Delta x, y + \Delta y, t + 1)) ~ +  \\
& \psi(\nabla I(x, y, t) - \nabla I(x + \Delta x, y + \Delta y, t + 1)) dxdy
\end{split}
\end{equation}
$\alpha$ denotes a weighting parameter and $\psi$ denotes a concave cost function, and $E_{smoothness}$ penalization term is introduced to avoid too significant motion displacement: 
\begin{equation}
    E_{smoothness} = \iint \psi(|\nabla x|^2 + |\nabla y|^2) dxdy
\end{equation}
In our approach, we utilize Liu's~\cite{celiu} dense optical flow to estimate motion over frame pairs.  However, while the intra-frame movement is estimable, it cannot be used directly as motion features because the content of each video varies considerably which makes the comparison between the estimated motion of different videos unreasonable~\cite{OFCNN}. Moreover, the estimated motion cannot be pixel-wise accurate due to the influence of noises and non-linear displacements.

\begin{figure}[t!]
    \centering
    
    \includegraphics[width=0.8\textwidth, height=0.38\textwidth]{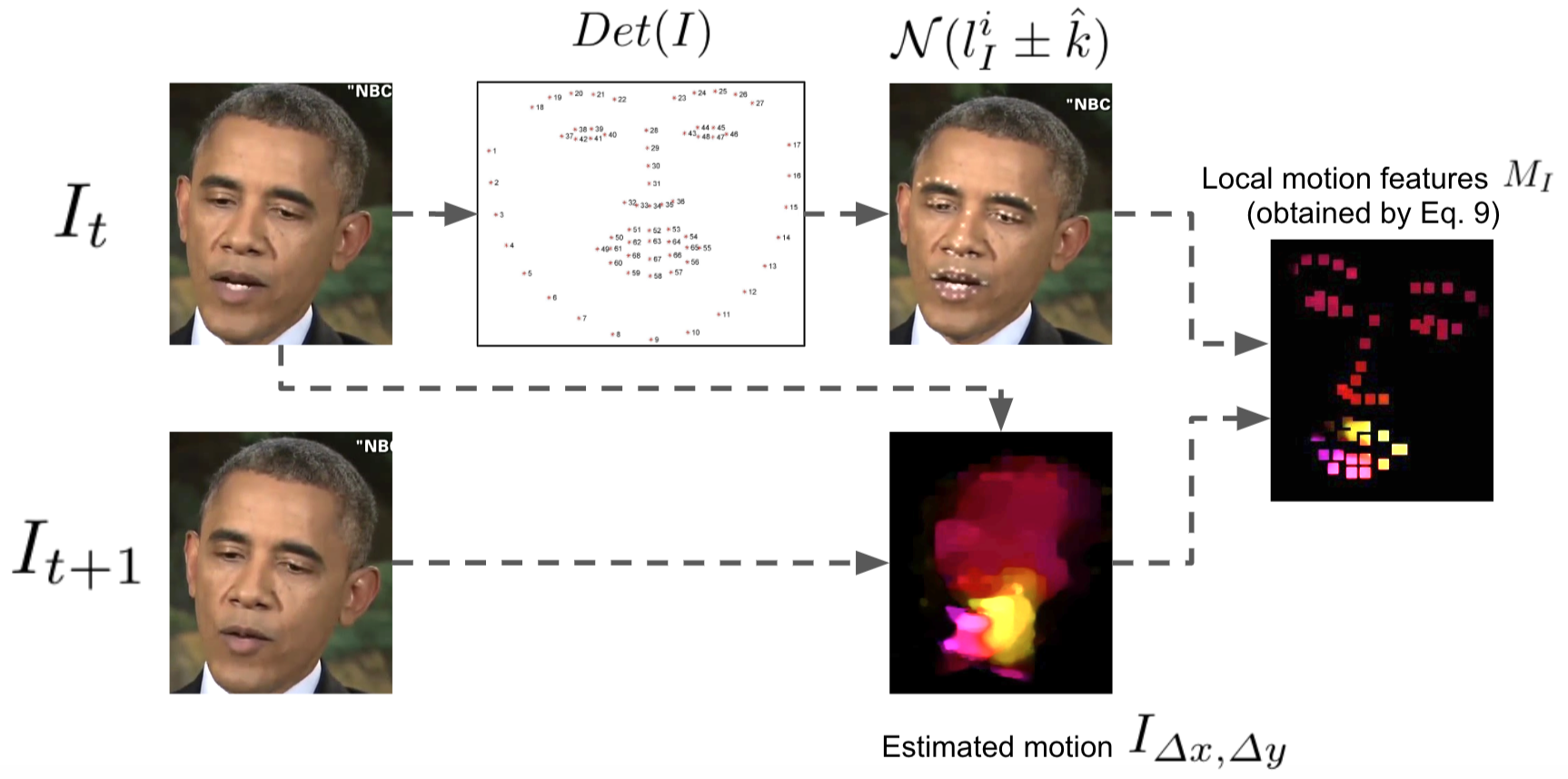}
    \caption{Illustration of local motion estimation step.}
    \label{fig:lme}
\end{figure}

To overcome the above problems, we propose to narrow the region of interests via finding facial landmarks for comparison. By employing an arbitrary facial landmark detector $f_{D}$, we are able to obtain a set of spatial coordinates $L$ as:
\begin{equation}
    f_D(I) = L_I = \{l^i_I | l_I^i \in \mathbb{R}^2, 1 \leq i \leq n \}
\end{equation}
so that the local motion features $M_I$ can be denoted as:
\begin{equation}
    M_I = \{m_I^i | m_I^i = I_{\Delta x, \Delta y} \oplus \mathcal{N}(l_I^i \pm \hat{k}), l_I^i \in L_I\}
\end{equation}
representing the Gaussian-weighted average of estimated motion map $I_{\Delta x, \Delta y}$ centered on $(l_i^x, l_i^y)$ with stride $\hat{k}$. The Gaussian smoothing is introduced to further mitigate the negative impact by inaccurate estimation result. By doing so, we align the motion feature extracted from each video for equitable comparison. An intuitive illustration of this step is presented in Fig.~\ref{fig:lme}. 
Due to the lack of sufficient motion in some $I_{\Delta x, \Delta y}$, we abandon these with trivial magnitude by setting a hyperparameter as threshold where the detailed choice will be discussed in Sect.~\ref{sect:Exp}. 

\subsection{Co-motion Patterns}
\label{sect:CMP}
Depending merely on local motion features obtained above would require an incredibly large-scale dataset to cover as many scenarios as possible, which is redundant and costly. Based on the observation that a human face is an articulated structure, the intra-component correlation can also depict the motion in an efficient manner. Inspired by the co-occurrence feature~\cite{Cooccurrence}, which has been frequently employed in texture analysis, we propose to further calculate the second-order statistics from extracted local motion features.

\subsubsection{Grouping Intra-Correlated Motion Features} \hfill \break
\noindent In this step, we group analogous $m_I^i \in M_I$ to estimate articulated facial structure by motion features since motion features that are collected from the same facial component would more likely to share consistent movement. Meanwhile, the negative correlation can also be represented where motion features having opposite directions (e.g. upper lip and lower lip) would be assigned to disjoint groups. 
As $m_I^i \in \mathbb{R}^2$ denotes motion on two orthogonal directions, we construct the affinity matrix $A_I$ on $M_I$ such that:
\begin{equation}
    A_I^{i, j} = m_I^i \cdot m_I^j
\end{equation}
We here choose the inner product over other metrics such as cosine and euclidean since we wish to both emphasize the correlation instead of difference and to lighten the impact of noise within $M_I$. In specific, using inner product can ensure the significance of two highly correlated motions that both possess certain magnitude to be highlighted, while noises with trivial magnitude would relatively affect less. The normalized spectral clustering~\cite{spectral,tutorial} is then performed, where we calculate the degree matrix $D$ such that:
\begin{equation}
    D_I^{i, j} = 
    \begin{cases}
      \sum^n_{j} A_I^{i, j} & \text{if $i = j$}\\
      0 & \text{if $i \neq j$}\\
    \end{cases}
\end{equation}
and the normalized Laplacian matrix $\mathcal{L}$ as:
\begin{equation}
    \mathcal{L} = (D_I)^{-\frac{1}{2}}(D_I - A_I)(D_I)^{\frac{1}{2}}
\end{equation}
In order to split $M_I$ into $K$ disjoint groups, the first $K$ eigenvectors of $\mathcal{L}$, denote as $\textbf{V} = \{\nu_k | k \in [1, K]\}$, are extracted to form matrix $F \in \mathbb{R}^{n \times K}$. After normalizing $F$ by dividing the corresponding L2-norms row-wisely, a K-Means clustering is used to separate $P = \{p_i | p_i = F^i \in \mathbb{R}^{K}, i \in [1, n]\}$ into $K$ clusters where $C_k = \{i | p_i \in C_k\}$. However, since $K$ is not directly available in our case, we will demonstrate how to determine the optimal $K$ in the next step.


\subsubsection{Constructing Co-motion Patterns} \hfill \break
As previously stated, determining a proper $K$ can also assist in describing the motion pattern more accurately. A straightforward approach is to iterate through all possible $K$ such that the Calinski-Harabasz index~\cite{CH} is maximized:
\begin{equation}
     \operatorname*{arg\,max}_{K \in [2, n]} ~ f_{CH}(\{C_k | k \in [1, K]\}, K)
\end{equation}
where
\begin{equation}
     f_{CH}(\{C_k | k \in [1, K]\}, K) = \frac{tr(\sum^K_y \sum_{p_i \in C_y} (p_i - C_y^{\mu})(p_i - C_y^{\mu})^\intercal)}{tr(\sum^K_y |C_y| (C_y^{\mu} - M_I^{\mu})(C_y^{\mu} - M_I^{\mu})^\intercal)} \times \frac{n - K}{K - 1}
\end{equation}
with $C_y^{\mu}$ is the centroid of $C_y$, $M_I^{\mu}$ is the center of all local motion features and $tr$ denotes taking the trace of the corresponding matrix. After all the efforts, the motion correlation matrix $\rho_{I_t, I_{t+1}}$ of two consecutive frames $I_t$ and $I_{t+1}$ can be calculated as:
\begin{equation}
    \rho_{I_t, I_{t+1}}^{i, j} = 
    \begin{cases}
      1 & \text{if $(m_i \in C_k ~\&~ m_j \in C_k ~|~  \exists C_k)$}\\
      0 & \text{otherwise}\\
    \end{cases}
\end{equation}
and consequently, the co-motion pattern of sequence $S = \{I_1, ..., I_T\}$ is calculated as the weighted average of all correlation matrices:
\begin{equation}
    f_{CP}(S) = \sum^T_t k_{I_t, I_{t+1}} \times f_{CH}(\{C_k | k \in [1, K]\}, k_{I_t, I_{t+1}}) \times \rho_{I_t, I_{t+1}}
\end{equation}
where the weighting procedure is also to reduce the impact of noise: the greater the $f_{CH}(\{C_k | k \in [1, K]\}, K)$, naturally the more consistent the motions are; simultaneously, co-motion pattern constructed on noisy estimated local motion would scatter more sparse, which should be weighted as less important.

\subsubsection{Usage of Co-motion Patterns} \hfill \break
The co-motion pattern can be utilized as a statistical feature for comparison purposes. When used for supervised classification, each co-motion must be normalized by its L1 norm:
\begin{equation}
    \dot f_{CP}(S) = \frac{f_{CP}(S)}{\sum |f_{CP}(S)|}
\end{equation}
and $\dot f_{CP}(S)$ can be used as features for arbitrary objectives. 
In order to illustrate that our co-motion pattern can effectively distinguish all forgery types by only modeling on real videos, we also conduct anomaly detection experiments where a real co-motion pattern is firstly built as template. Then, co-motion patterns from real and forgery databases are all compared against the template where the naturalness is determined by the threshold. 
Jensen–Shannon divergence is suggested to be employed as distance measure between any two co-motion patterns:
\begin{equation}
    d_{KL}(f_{CP}(S_1), f_{CP}(S_2)) = \sum_i \sum_j^{i-1} f_{CP}(S_1)^{i, j} log(\frac{f_{CP}(S_1)^{i, j}}{f_{CP}(S_2)^{i, j}}) 
\end{equation}
\begin{equation}
    d_{JS}(f_{CP}(S_1), f_{CP}(S_2)) = \frac{1}{2} d_{KL}(f_{CP}(S_1), \overline{f_{CP}}_{S_1, S_2}) + \frac{1}{2} d_{KL}(f_{CP}(S_2), \overline{f_{CP}}_{S_1, S_2})
\end{equation}
where $\overline{f_{CP}}_{S_1, S_2} = \frac{f_{CP}(S_1) + f_{CP}(S_2)}{2}$ and $S1, S2$ denote two sequences.

%% file: Sec_Exp.tex
\section{Experiments}
\label{sect:Exp}
In this section, extensive experiments are conducted to empirically demonstrate the feasibility of our co-motion pattern, coupled with the advantages over other methods. We first describe the experiment protocol, followed by the choice of hyperparameters. The quantitative performance of our method evaluated on different datasets is reported and analyzed in Sect.~\ref{sec:quantitative}. Subsequently, we interpret the composition of the co-motion pattern, showing how it can be used for determining the genuineness of any given sequence or even individual estimated motion set. Finally, we demonstrate the transferability and robustness of our method under different scenarios. 

\subsubsection{Dataset}
We evaluate our method on FaceForensics++~\cite{FaceForensics} dataset which consists of four sub-databases that produce face forgery via different methods, i.e. Deepfake~\cite{deepfake}, FaceSwap~\cite{faceswap}, Face2Face~\cite{F2F} and NeuralTexture~\cite{NeuralTexture}. In addition, we utilize the real set from~\cite{Google_dataset} to demonstrate the similarity of co-motion patterns from real videos.
Since each sub-database contains 1,000 videos, we form 2,000 co-motion patterns with each composed of picking $N$ $\rho$ matrices for training and testing respectively.  
We use c23 and c40 to indicate the quality of datasets, which are compressed by H.264~\cite{H264} with 23 and 40 as constant rate quantization parameters. 
Unless otherwise stated, all of our performance reported are achieved on c23. 
The validation set and testing set are split before any experiments to ensure no overlapping would interfere the results. 

\subsubsection{Implementation}
In this section, we specify hyperparameters and other detailed settings in order to reproduce our method. The local motion estimation procedure is accomplished by integrating  \cite{opticalflow} as the estimator and \cite{Landmark} as the landmark detector, both with default parameter settings as reported in the original papers. For the facial landmarks, we only keep the last 51 landmarks out of 68 in total as the first 17 denotes the face boundary which is usually not manipulated. During the calculation of co-motion, we constrain $K$ to be at most 8 as only 8 facial components, thus avoiding unnecessary computation. 
Since a certain portion of frames do not contain sufficient motion, we only preserve co-motion patterns with $p\%$ motion features having greater magnitude than the total $p\%$ of others, i.e. $p = 0.5$ with magnitude $\geq 0.85$, where the number is acquired by randomly sampling a set of 100 videos. An AdaBoost~\cite{AdaBoost} classifier is employed for all supervised classification tasks. 
For Gaussian smoothing, we set $\hat{k} = 3$ for all experiments. 

\subsection{Quantitative Results}
\label{sec:quantitative}

\begin{table}[t!]
    \caption{Accuracy of our method on all four forgery databases, with each treated as a binary classification task against the real videos. Performance of  \cite{OpenWorld} is estimated from figures in the paper.
    }
    \begin{center}
    \begin{tabular}{l|c|c|c|c|c}
    \hline
    Method/Dataset                                 & Deepfakes & FaceSwap & Face2Face & NeuralTexture & Combined \\ \hline
    Xception~\cite{FaceForensics} & 93.46\%   & 92.72\%  & 89.80\%    & N/A            & \textbf{95.73\%}  \\
    R-CNN~\cite{RCNN}            & 96.90\%    & 96.30\%   & \textbf{94.35\%}   & N/A            & N/A       \\
    Optical Flow + CNN~\cite{OFCNN}           & N/A        & N/A       & 81.61\%   & N/A            & N/A       \\
    FacenetLSTM~\cite{OpenWorld}       & 89\%      & 90\%     & 87\%      & N/A            & N/A       \\ \hline
    $N$ = 1 (Ours)                             & 63.65\%   & 61.90\%  & 56.50\%   & 56.65\%       & 57.05\%        \\
    $N$ = 10 (Ours)                           & 82.80\%   & 81.95\%  & 72.30\%   & 68.50\%       & 71.30\%        \\
    $N$ = 35  (Ours)                          & 95.95\%   & 93.60\%  & 85.35\%   & 83.00\%       & 88.25\%        \\
    $N$ = 70  (Ours)                          & \textbf{99.10\%}   & \textbf{98.30\%}  & 93.25\%   & \textbf{90.45\%}       & 94.55\%      \\ \hline
    \end{tabular}
    \end{center}
\end{table}

In this section, we demonstrate the quantitative results of our method under different settings. At first, we show that the co-motion pattern can adequately separate forged and real videos in classification tasks as shown in Tab.~1. Comparing with other state-of-the-art forensic methods in terms of classification accuracy, we have achieved competent performance and have outperformed them by a large margin on Deepfakes~\cite{deepfake} and FaceSwap~\cite{faceswap}, respectively $99.10\%$ and $98.30\%$. In \cite{OFCNN}, while the researchers have similarly attempted establishing a forensic pipeline on top of motion features, we have outperformed its performance by approx. 12$\%$. It is noteworthy that \cite{RCNN,OpenWorld,FaceForensics} are all exploiting deep features that are learned in an end-to-end manner and consequently cannot be properly explained. By contrast, as interpretability is one of the principal factors to media forensics, our attention lies on proposing a method such that it can be justified and make no effort on deliberately outperforming deep learning based methods.  

\begin{figure}[t]
        \centering
        \begin{subfigure}[b]{0.485\textwidth}
            \centering
            \includegraphics[width=\textwidth]{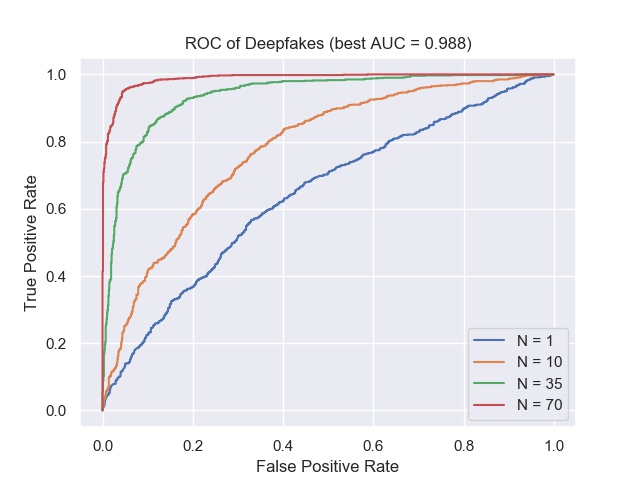}
        \end{subfigure}
        \hfill
        \begin{subfigure}[b]{0.485\textwidth}  
            \centering 
            \includegraphics[width=\textwidth]{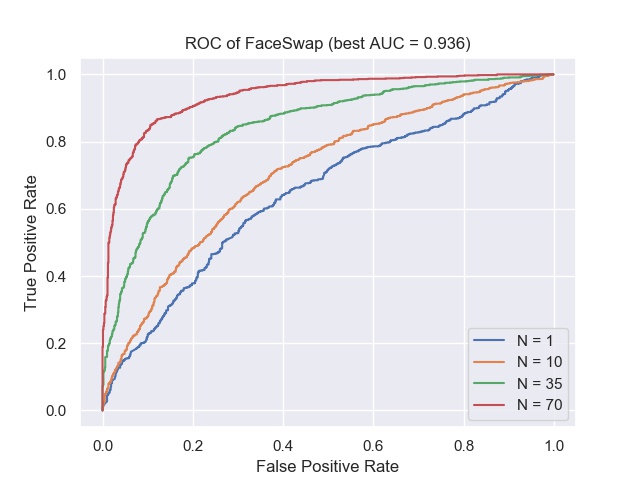}
        \end{subfigure}
        \vskip\baselineskip
        \begin{subfigure}[b]{0.485\textwidth}   
            \centering 
            \includegraphics[width=\textwidth]{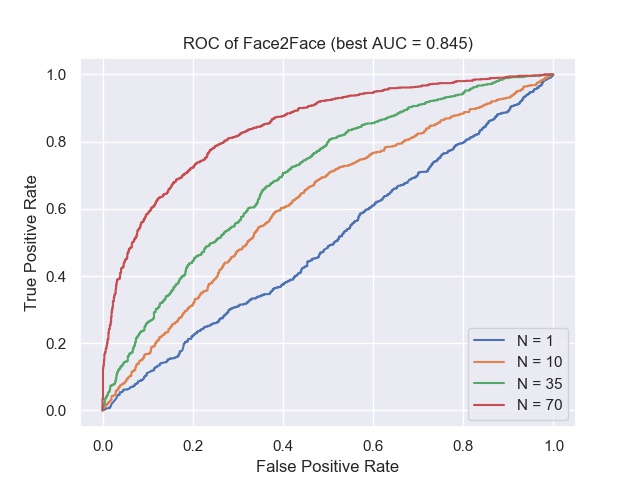}

        \end{subfigure}
        \hfill
        \begin{subfigure}[b]{0.485\textwidth}   
            \centering 
            \includegraphics[width=\textwidth]{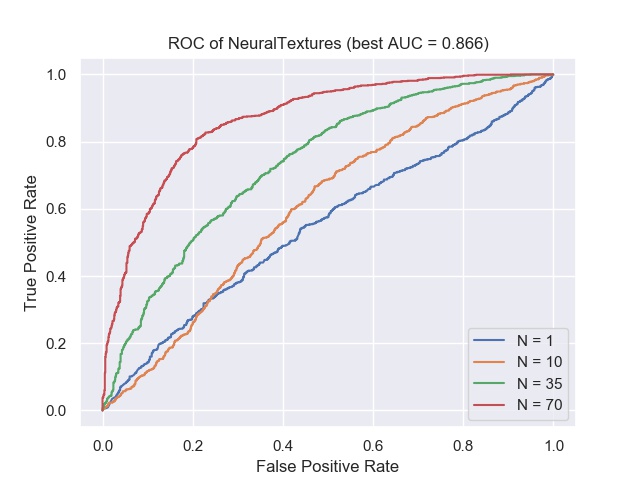}

        \end{subfigure}
        \caption{Anomaly detection performance of our co-motion patterns. }
        \label{fig:ROCs}
    \end{figure}

Equally importantly, as forgery methods are various and targeting each is expensive, we demonstrate that the proposed co-motion pattern can also be employed for anomaly detection tasks, where only the behaviors of real videos require to be modeled, and forged videos can be separated if an appropriate threshold is selected. As presented in Fig.~\ref{fig:ROCs}, we show receiver operating characteristic (ROC) curves on each forgery database with increasing $N$. The real co-motion template is constructed of 3,000 randomly selected $\rho$ matrices for each co-motion pattern (real or fake) to compare against during evaluation. In general, our method can be used for authenticating videos even without supervision. In the next section, we exhibit that the co-motion pattern is also robust to random noise and data compression. 

\vspace{-0.3cm}
\subsection{Robustness Analysis}
\label{sec:robustness}
In this section, we demonstrate the robustness of our proposed method against noises or data compression and the generalizability of co-motion patterns. Experiments about whether the compression rate of the video and noise would affect the effectiveness of co-motion patterns are conducted and the results are shown in Tab. 2. Empirically, co-motion has demonstrated great robustness against heavy compression (c40) and random noise, i.e. $N(\mu,\sigma^2)$ with $\mu = 0$ and $\sigma = 1$. Such results verify our proposed co-motion patterns exploiting high-level temporal information are much less sensitive to pixel-level variation, while statistics based methods as reviewed in Sect.~2.2 do not possess this property.
\vspace{-0.7cm}
\begin{table}
\caption{Robustness experiment for demonstrating that co-motion can maintain its characteristics under different scenarios. All experiments are conducted on Deepfake~\cite{deepfake} with $N = 35$. Classification accuracy and area under curve (AUC) are reported respectively. }
\begin{center}
    \begin{tabular}{l|c|c|c|c}
    \hline
    Setting / Dataset        & Original &c23&c40& c23+noise \\ \hline
    Binary classification    & 97.80\%        & 95.95\%                & 91.60\%               & 91.95\%                    \\ \hline
    Anomaly detection & 98.57        & 96.14                & 93.76               & 92.60                    \\ \hline
    \end{tabular}
\end{center}
\end{table}

\vspace{-0.5cm}
In addition to demonstrating the robustness, we also investigate in whether the modeled co-motion patterns are generalizable, as recorded in Tab.~3. It turns out that co-motion patterns constructed on relatively high-quality forgery databases such as NeuralTextures~\cite{NeuralTexture} and Face2Face~\cite{F2F} can easily be generalized for classifying other low-quality databases, while the opposite results in inferior accuracy.  This phenomenon is caused by that videos forged by NeuralTextures are generally being more consistent, thus the inconsistency learned is more narrowed down and specific, while the types of inconsistency vary greatly in low-quality databases, which can be hard to model.

\vspace{-0.5cm}
\begin{table}

\caption{Experiments for demonstrating generalizability of co-motion patterns. Same experiment setting was employed as in Tab. 1. }
\begin{center}
    \begin{tabular}{l|c|c|c|c}
    \hline Test on / Train on & Deepfakes & FaceSwap & Face2Face & NeuralTexture \\ \hline
    Deepfakes          & N/A         & 92.15\%        & 93.45\%         & 95.85\%            \\
    FaceSwap           & 84.25\%         & N/A        & 76.75\%         & 84.95\%             \\
    Face2Face          & 70.30\%         & 64.85\%        & N/A         & 81.65\%             \\
    NeuralTexture      & 76.20\%          & 65.15\%        & 77.85\%           & N/A           \\ \hline
    \end{tabular}
\end{center}
\end{table}

\vspace{-1cm}
\subsection{Abnormality Reasoning}
\label{sec:reasoning}
In this section, we explicitly interpret the implication of each co-motion pattern for an intuitive understanding. A co-motion example of real videos can be found in Fig. ~6. As we illustrated, the local motion at 51 facial landmarks are estimated as features, where the order of landmarks are preserved identically in all places on purpose for better visual understanding. It is noteworthy that the order of landmarks do not affect the performance as long as they are aligned during experiments. 

Consequently, each co-motion pattern describes the relationship of any pair of two local motion features, where features from the same or highly correlated facial component would instinctively have greater correlation. For instance, it is apparent that two eyes would generally move in the same direction, as the center area highlighted in Fig. ~6. Similarly, a weak yet stable high correlation of the first 31 features is consistently observed on all real co-motion patterns, which conforms to the accordant movement of facial components on upper and middle face area. We also observe strong negative correlation, indicating opposite movements, between upper lip and lower lip. This credits to the dataset containing a large volume of videos with people talking, while in forged videos such a negative correlation is undermined, usually due to the fact that the videos are synthesized in a frame-by-frame manner, thus the temporal relationship is not well-preserved. Moreover, the co-motion is normalized in range $[0, 1]$ for visualization purpose which leads to the weakened difference between real and fake co-motion patterns, while in original scale the difference can be more magnificent, verified by the experiments.

\begin{figure}[t!]
    \centering
    \includegraphics[width=0.55\textwidth, height=0.48\textwidth]{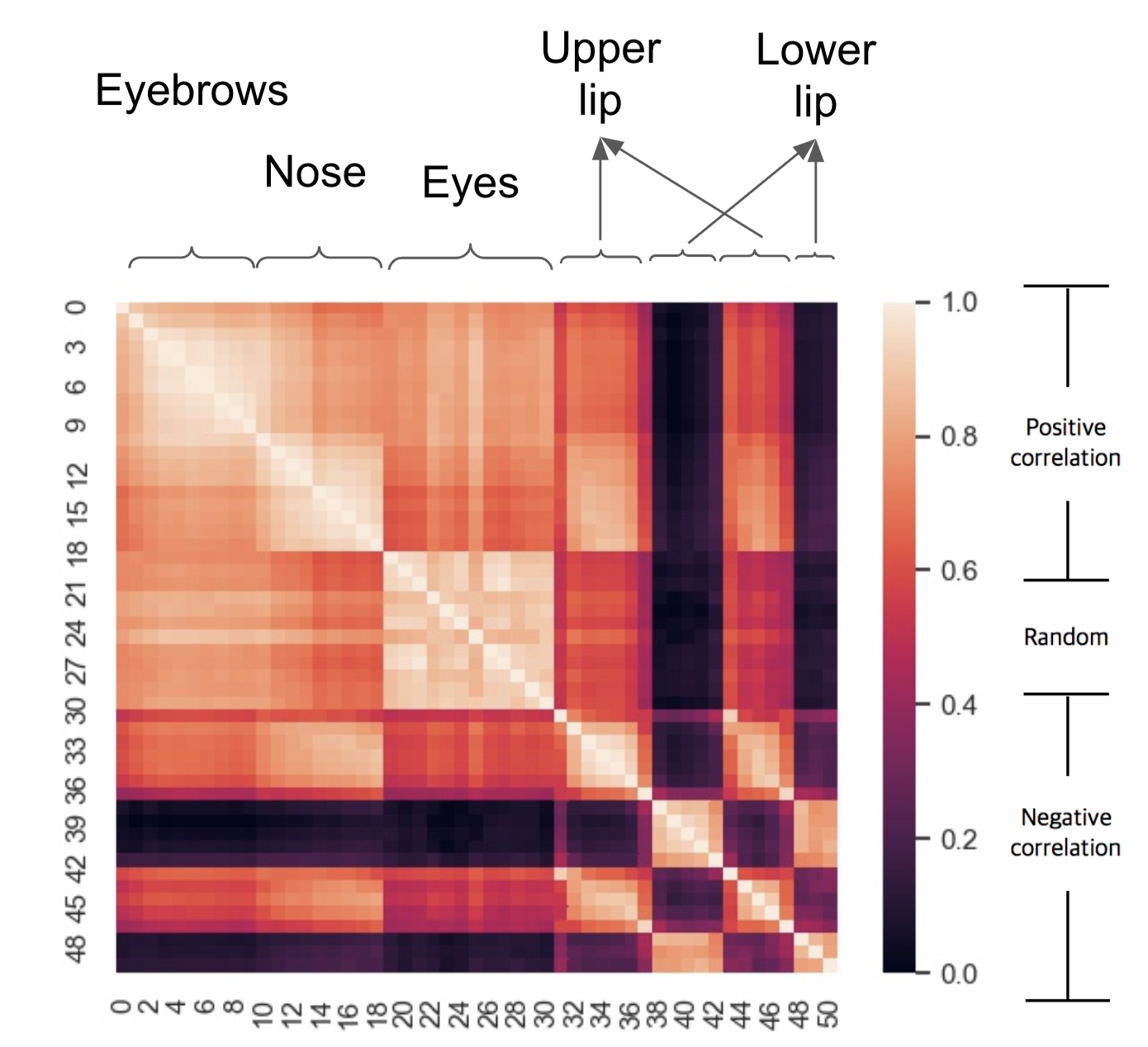}
    \caption{An example of interpreting co-motion patterns. }
    \label{fig:interpret}
\end{figure}

For an explicit comparison, we also average 1,000 $\rho$ matrices from each source to illustrate the distinction and which motion pattern in specific was not well-learned as in Fig.~7. Evidently, co-motion patterns from forged videos fail to model the negative correlation between upper lip and lower lip. Moreover, in Deepfake and FaceSwap, the positive correlation between homogeneous components (e.g. eyes and eyebrows) is also diluted, while in reality it would be difficult to control them having uncorrelated motion. We also attempt to construct co-motion patterns on another set of real videos~\cite{Google_dataset} to illustrate the commonality of co-motion patterns over all real videos. Additionally, we show that visually, the structure of co-motion pattern could quickly converge as illustrated in Fig. 8, which sustains our choices of building second-order pattern as it is less sensitive to intra-instance variation. 

\begin{figure*}[t]
        \centering
        \begin{subfigure}[b]{0.3\textwidth}
            \centering
            \includegraphics[width=0.83\textwidth]{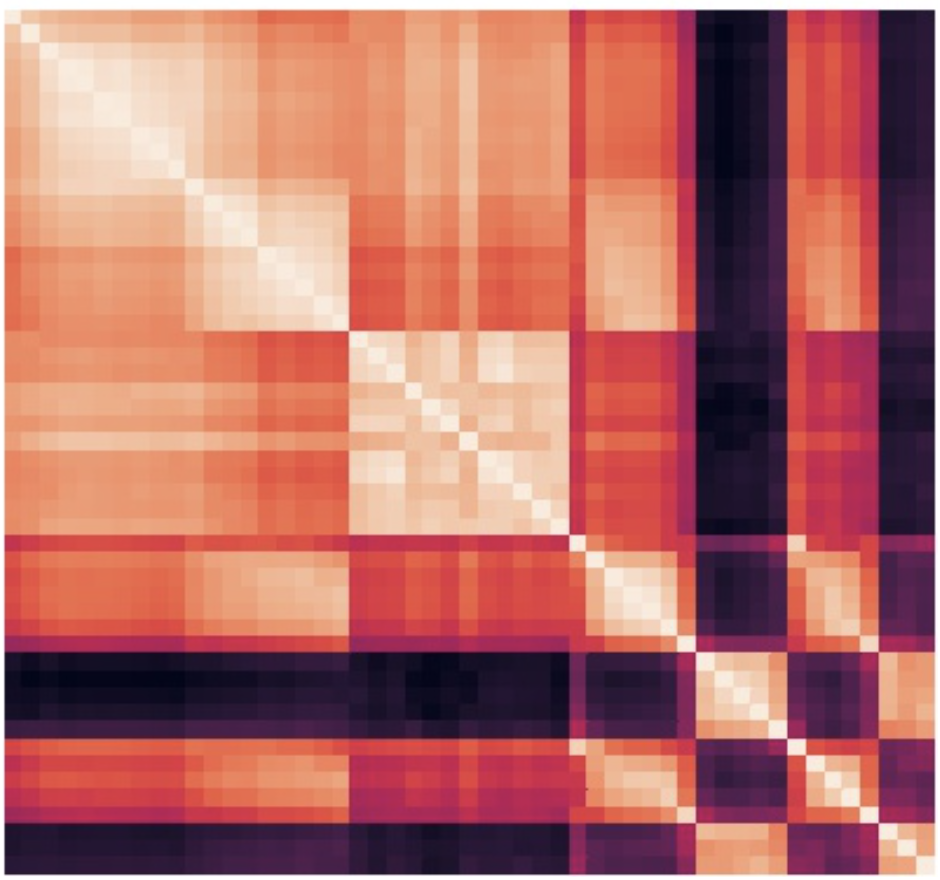}
            {{\small Real videos}}    
        \end{subfigure}
        \hfill
        \begin{subfigure}[b]{0.3\textwidth}  
            \centering 
            \includegraphics[width=0.83\textwidth]{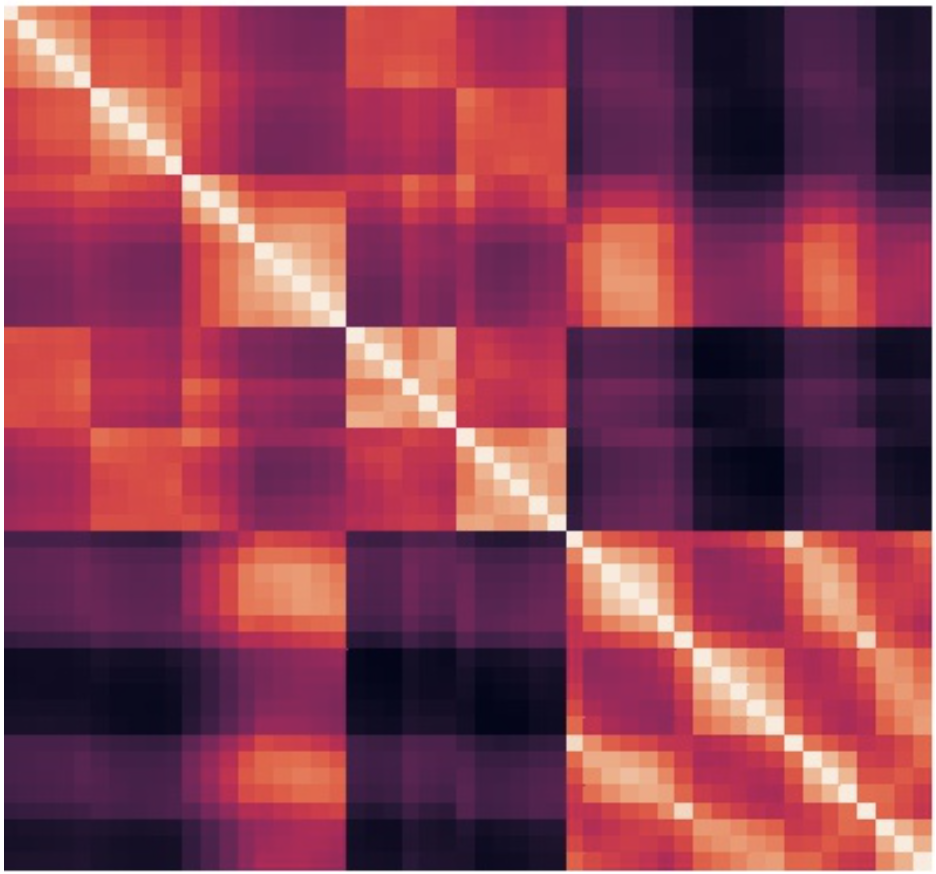}
            {{\small Deepfakes}}    
        \end{subfigure}
        \hfill
        \begin{subfigure}[b]{0.3\textwidth}  
            \centering 
            \includegraphics[width=0.83\textwidth]{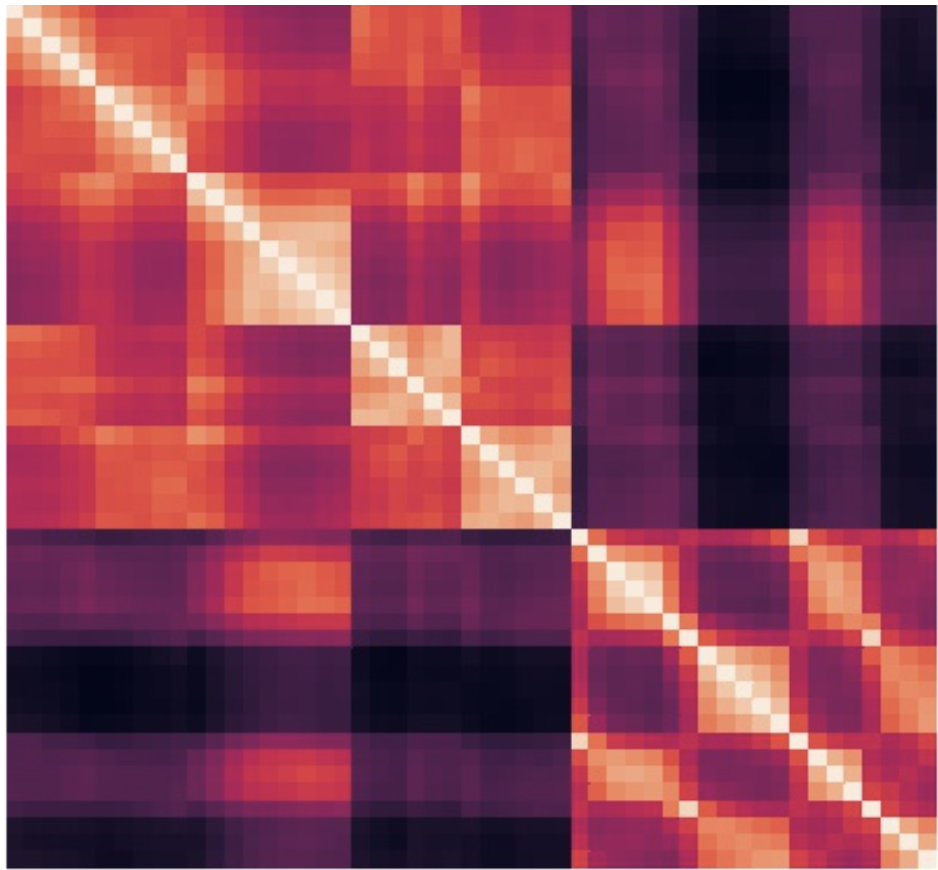}
            {{\small FaceSwap}}    
        \end{subfigure}
        \vskip\baselineskip
        \begin{subfigure}[b]{0.3\textwidth}   
            \centering 
            \includegraphics[width=0.83\textwidth]{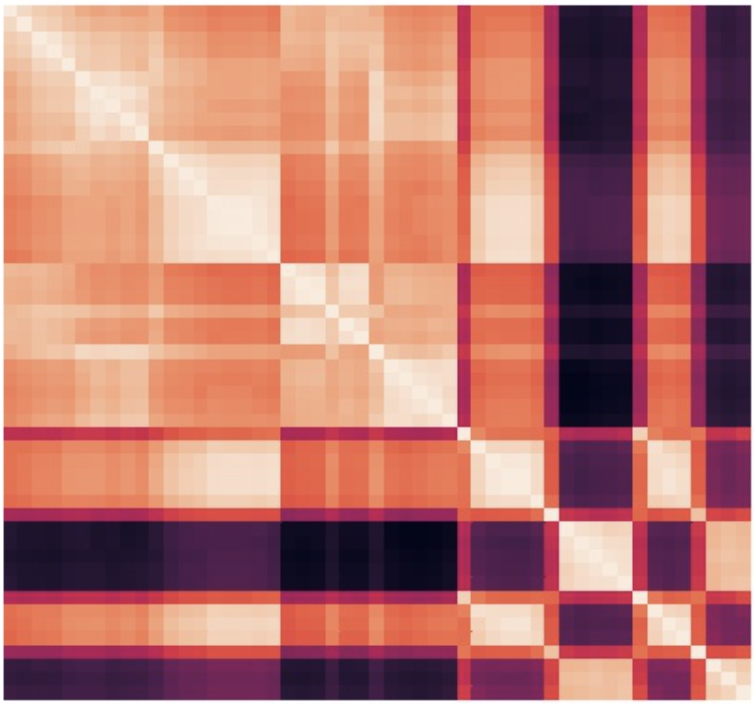}
            {{\small Real videos from \cite{Google_dataset}}}    
        \end{subfigure}
        \hfill
        \begin{subfigure}[b]{0.3\textwidth}   
            \centering 
            \includegraphics[width=0.83\textwidth]{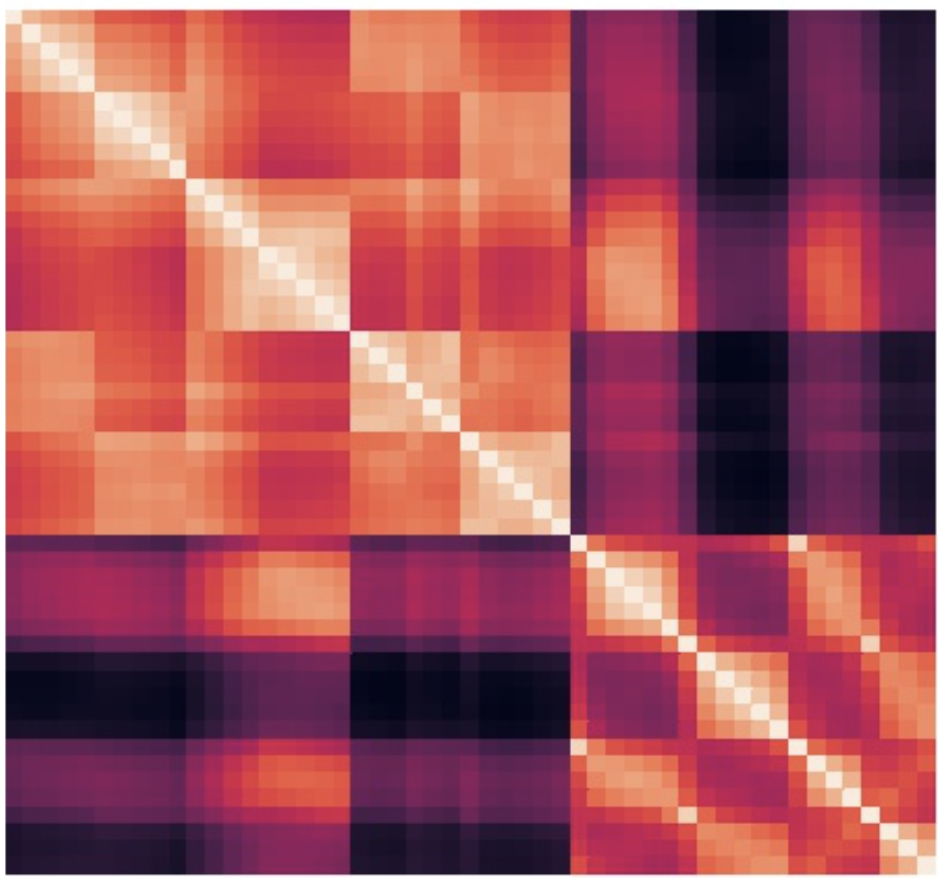}
            {{\small Face2Face}}    
            \label{fig:mean and std of net44}
        \end{subfigure}
        \hfill
        \begin{subfigure}[b]{0.3\textwidth}  
            \centering 
            \includegraphics[width=0.83\textwidth]{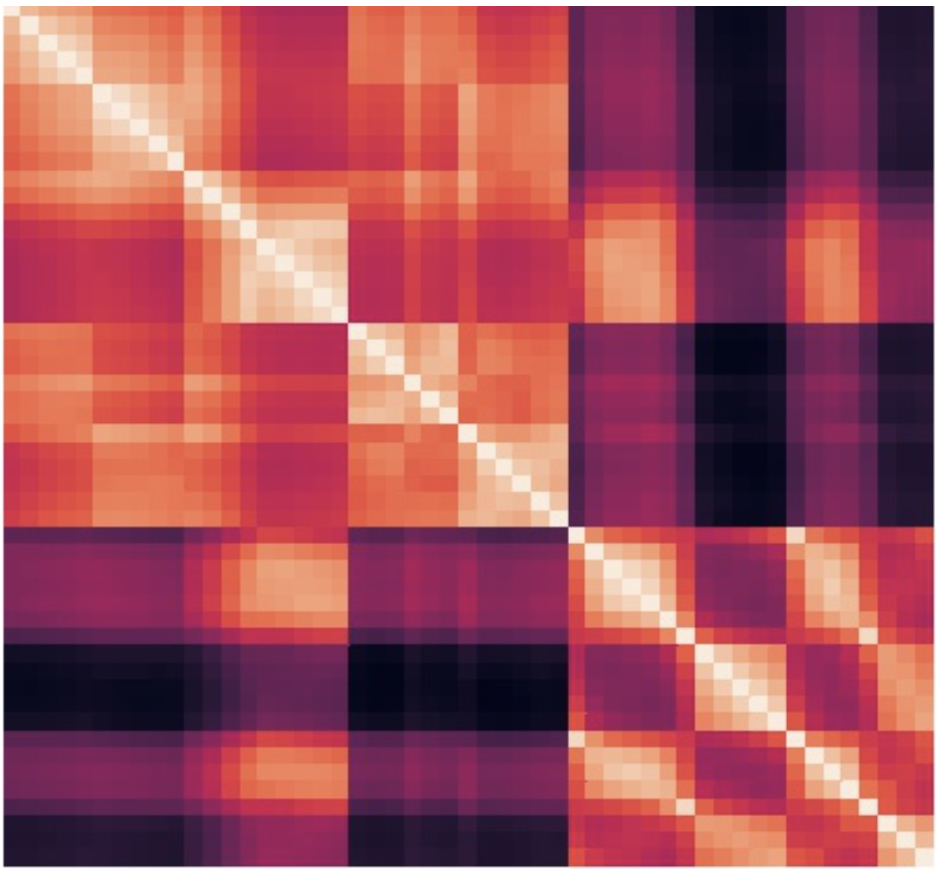}
            {{\small NeuralTexture}}    
        \end{subfigure}
       \caption{Averaged co-motion pattern from different sources. Two real co-motion patterns (leftmost column) collectively present component-wise motion consistency while forged videos fail to maintain that property. }
        \label{fig:Cooccurrences}
    \end{figure*}

\begin{figure}[t!]
    \centering
    \includegraphics[width=0.68\textwidth, height=0.3\textwidth]{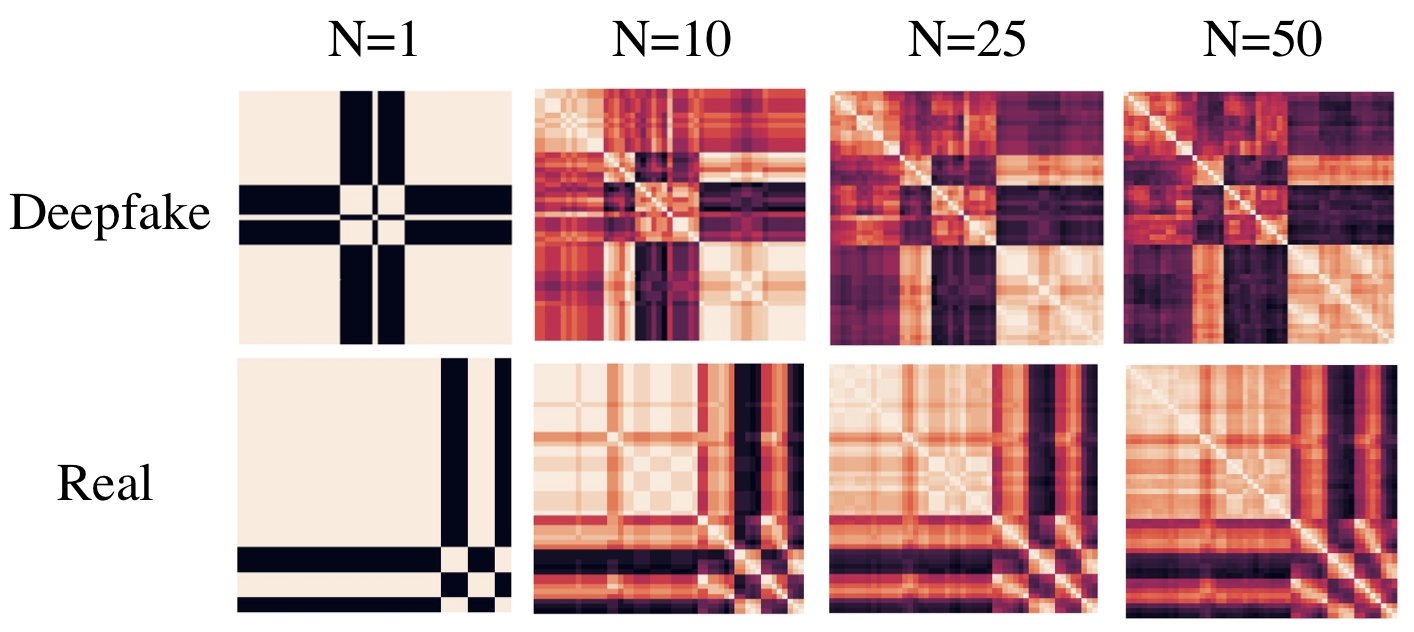}
    \caption{Co-motion pattern comparison on the same video (original and deep-faked based on the original one). As $N$ increases, both co-motion patterns gradually converge to the same structure. }
    \label{fig:framework}
\end{figure}